\documentclass[11pt,a4paper]{article}
\usepackage[hyperref]{acl2020}

\usepackage{times}
\usepackage{latexsym}
\usepackage{microtype}
\usepackage{natbib}
\usepackage{booktabs}
\usepackage{appendix}
\usepackage{tikz}
\usepackage{tikz-dependency}
\usepackage{amsmath}
\usepackage{amsfonts}
\usepackage{graphicx}
\usepackage{amsthm}
\usepackage{mathtools}
\usepackage{multirow}
\usepackage{url}
\usepackage{color}
\usepackage{subfig}
\usepackage{dsfont}
\usepackage[ruled,vlined,noend]{algorithm2e}

\usepackage[noabbrev,capitalize]{cleveref} 

\crefname{equation}{equation}{equations}   
\crefname{footnote}{footnote}{footnotes}   
\crefname{line}{line}{lines}               
\crefname{section}{\S}{\S\S}
\Crefname{section}{\S}{\S\S}    
\crefformat{section}{#2\S#1#3}  
\Crefformat{section}{#2\S#1#3}
\crefrangeformat{section}{\S\S#3#1#4--#5#2#6}
\Crefrangeformat{section}{\S\S#3#1#4--#5#2#6}

\newcommand{\exper}[1]{\textsc{#1}}

\newcommand{\LM}{\exper{LM}}

\newcommand{\prefix}{\exper{Prefix}}
\newcommand{\adapter}{\exper{Adapter}}
\newcommand{\FTtop}{\exper{FT-top2}}
\newcommand{\Emb}{\exper{Emb}}
\newcommand{\FT}{\exper{Fine-tune}}
\newcommand{\infix}{\exper{Infix}}
\newcommand{\MLP}{\exper{MLP}}
\newcommand{\xx}{\textsf{X$_{\text{idx}}$}}
\newcommand{\yy}{\textsf{Y$_{\text{idx}}$}}
\newcommand{\pp}{\textsf{P$_{\text{idx}}$}}

\newcommand{\PrefixMatrix}{P}

\DeclareMathOperator{\Softmax}{softmax}

\usepackage{microtype}

\aclfinalcopy 

\title{Prefix-Tuning: Optimizing Continuous Prompts for Generation}

\author{Xiang Lisa Li \\
  Stanford University \\
  \texttt{xlisali@stanford.edu} \\\And
  Percy Liang \\
  Stanford University \\
  \texttt{pliang@cs.stanford.edu} \\}

\date{}

\begin{document}
\maketitle
\begin{abstract}
Fine-tuning is the de facto way to leverage large pretrained language models to perform downstream tasks. 
However, it modifies all the language model parameters and therefore necessitates storing a full copy for each task.
In this paper, we propose prefix-tuning, a lightweight alternative to fine-tuning for natural language generation tasks, which keeps language model parameters frozen, but optimizes a small \emph{continuous task-specific} vector (called the prefix). 
Prefix-tuning draws inspiration from prompting, allowing subsequent tokens to attend to this prefix as if it were ``virtual tokens''. 
We apply prefix-tuning to GPT-2 for table-to-text generation and to BART for summarization. 
We find that by learning only 0.1\% of the parameters, prefix-tuning obtains comparable performance in the full data setting, outperforms fine-tuning in low-data settings, and extrapolates better to examples with topics unseen during training.
\end{abstract}

\section{Introduction}
\label{sec:intro}

Fine-tuning is the prevalent paradigm for using large pretrained language models (LMs) \cite{Radford2019gpt2,devlin-etal-2019-bert} to perform downstream tasks (e.g., summarization), but it requires updating and storing all the parameters of the LM. 
Consequently, to build and deploy NLP systems that rely on large pretrained LMs, one currently needs to store a modified copy of the LM parameters for each task.  This can be prohibitively expensive, given the large size of current LMs; for example, GPT-2 has 774M parameters \cite{Radford2019gpt2} and GPT-3 has 175B parameters \cite{brown2020language}. 

\begin{figure}
    \centering
    \includegraphics[page=1,width=0.48\textwidth]{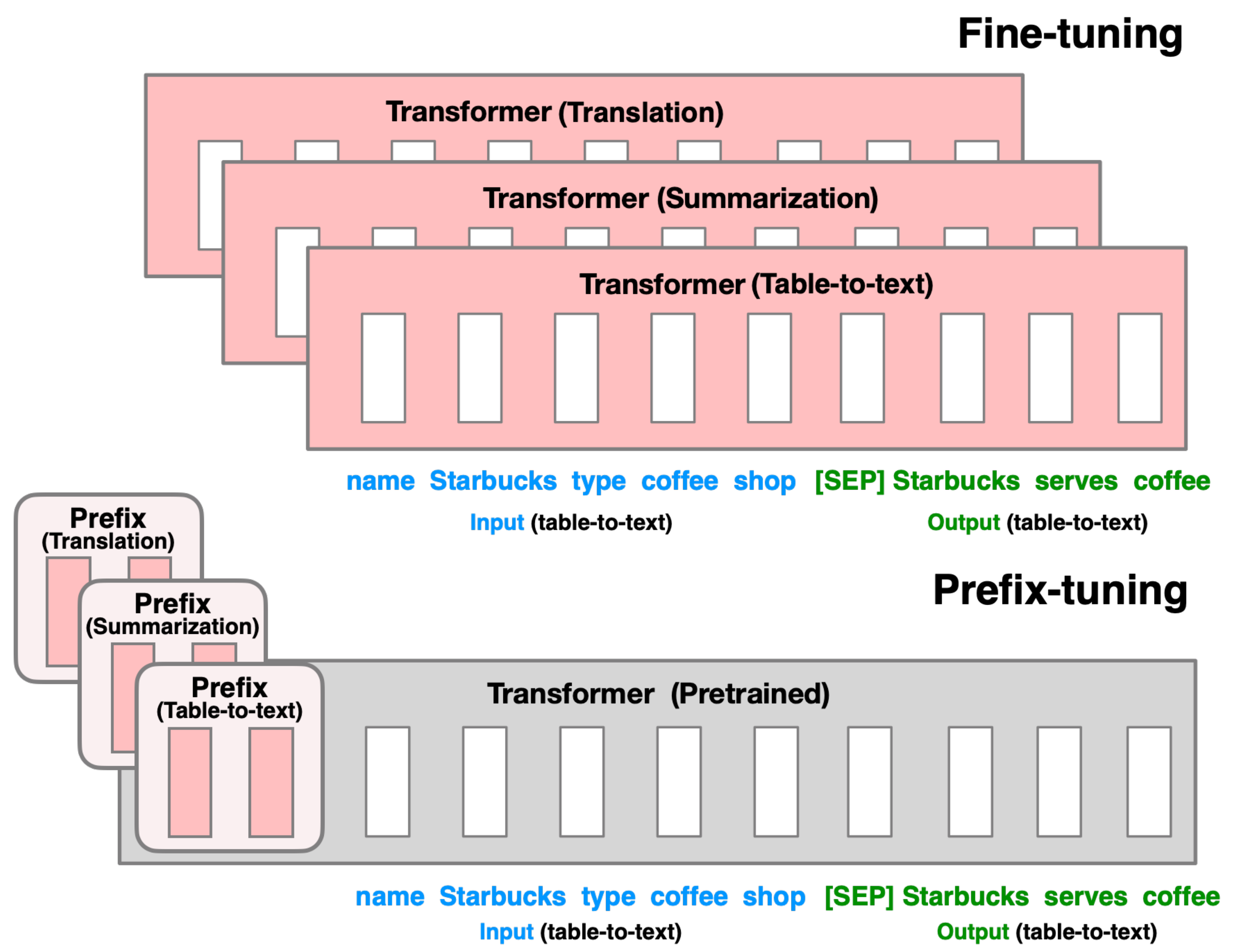}
    \vspace{-23pt}
    \caption{\label{fig:1} Fine-tuning (top) updates all Transformer parameters (the red Transformer box) and requires storing a full model copy for each task. 
    We propose prefix-tuning (bottom), which freezes the Transformer parameters and only optimizes the prefix (the red prefix blocks). Consequently, we only need to store the prefix for each task, making prefix-tuning modular and space-efficient. Note that each vertical block denote transformer activations at one time step.
    }
    \vspace{-19pt}
\end{figure}

A natural approach to this problem is \emph{lightweight fine-tuning}, which freezes most of the pretrained parameters and augments the model with small trainable modules.
For example, adapter-tuning \cite{Rebuffi2017Adapter,pmlr-v97-houlsby19a} inserts additional task-specific layers between the layers of pretrained language models.
Adapter-tuning has promising performance on natural language understanding and generation benchmarks, attaining comparable performance with fine-tuning while adding only around 2-4\% task-specific parameters \cite{pmlr-v97-houlsby19a,lin-etal-2020-exploring}.

On the extreme end, GPT-3 \cite{brown2020language} can be deployed without any task-specific tuning. 
Instead, users prepend a natural language task instruction (e.g., \textit{TL;DR} for summarization) and a few examples to the task input; then generate the output from the LM. This approach is known as in-context learning or \emph{prompting}.

In this paper, we propose \textit{prefix-tuning}, a lightweight alternative to fine-tuning for natural language generation (NLG) tasks, inspired by prompting. 
Consider the task of generating a textual description of a data table, as shown in \cref{fig:1}, where the task input is a linearized table (e.g., ``name: Starbucks $|$ type: coffee shop'') and the output is a textual description (e.g., ``Starbucks serves coffee.''). 
Prefix-tuning prepends a sequence of  \emph{continuous} \emph{task-specific} vectors to the input, which we call a  \emph{prefix}, depicted by red blocks in \cref{fig:1} (bottom).  For subsequent tokens, the Transformer can attend to the prefix as if it were a sequence of ``virtual tokens'', but unlike prompting, the prefix consists entirely  of free parameters which do not correspond to real tokens.
In contrast to fine-tuning in \cref{fig:1} (top), which updates all Transformer 
parameters and thus requires storing a tuned copy of the model for each task, 
prefix-tuning only optimizes the prefix. 
Consequently, we only need to store one copy of the large Transformer and a learned task-specific prefix, yielding a very small overhead for each additional task (e.g., 250K parameters for table-to-text).

In contrast to fine-tuning, prefix-tuning is modular: we train an upstream prefix which steers a downstream LM, which remains unmodified.
Thus, a single LM can support many tasks at once.
In the context of personalization where the tasks correspond to different users \cite{Shokri-Shmatikov-2015, McMahan-et-al-2016},
we could have a separate prefix for each user trained only on that user's data,
thereby avoiding data cross-contamination.
Moreover, the prefix-based architecture enables us to even process examples from multiple users/tasks in a single batch,
something that is not possible with other lightweight fine-tuning approaches.

We evaluate prefix-tuning on table-to-text generation using GPT-2 and abstractive summarization using BART. In terms of storage, prefix-tuning  stores 1000x fewer parameters than fine-tuning. In terms of performance when trained on full datasets, prefix-tuning and fine-tuning are comparable for table-to-text (\cref{ssec:table-to-text}), while prefix-tuning suffers a small degradation for summarization (\cref{ssec:summarization}). In low-data settings, prefix-tuning on average outperforms fine-tuning on both tasks (\cref{sec:lowdata}). 
Prefix-tuning also extrapolates better to tables (for table-to-text) and articles (for summarization) with unseen topics (\cref{sec:extrapolation}). 

\section{Related Work}


\label{sec:related_work}
\paragraph{Fine-tuning for natural language generation.} 
Current state-of-the-art systems for natural language generation are based on fine-tuning pretrained LMs. 
For table-to-text generation, \citet{kale2020texttotext} fine-tunes a sequence-to-sequence model \citep[T5;][]{T5}. 
For extractive and abstractive summarization, researchers fine-tune masked language models \citep[e.g., BERT;][]{devlin-etal-2019-bert} and encode-decoder models \citep[e.g., BART;][]{lewis-etal-2020-bart}  respectively \cite{zhong-etal-2020-extractive, liu-lapata-2019-text,T5}.
For other conditional NLG tasks such as machine translation and dialogue generation, fine-tuning is also the prevalent paradigm \cite{zhang-etal-2020-dialogpt, stickland2020recipes, Zhu2020Incorporating, liu2020multilingual}. 
In this paper, we focus on table-to-text using GPT-2 and summarization using BART, but prefix-tuning can be applied to other generation tasks and pretrained models.

\paragraph{Lightweight fine-tuning.}
Lightweight fine-tuning freezes most of the pretrained parameters and modifies the pretrained model with small trainable modules. The key challenge is to identify high-performing architectures of the modules and the subset of pretrained parameters to tune. 
One line of research considers removing parameters: some model weights are ablated away by training a binary mask over model parameters \cite{zhao2020masking,dixit2020How}.
Another line of research considers inserting parameters.
For example, \citet{zhang2020sidetuning} trains a ``side'' network that is fused with the pretrained model via summation; adapter-tuning inserts task-specific layers (adapters) between each layer of the pretrained LM \cite{pmlr-v97-houlsby19a,lin-etal-2020-exploring,Rebuffi2017Adapter,pfeiffer2020adapterfusion}.
Compared to this line of work, which tunes around $3.6\%$ of the LM parameters, our method obtains a further 30x reduction in task-specific parameters, tuning only 0.1\% while maintaining comparable performance. \looseness=-1

\paragraph{Prompting.} 

Prompting means prepending instructions and a few examples to the task input and generating the output from the LM. GPT-3 \cite{brown2020language} uses manually designed prompts to adapt its generation for different tasks, and this framework is termed \emph{in-context learning}. However, since Transformers can only condition on a bounded-length context (e.g., 2048 tokens for GPT-3), in-context learning is unable to fully exploit training sets longer than the context window. 
\citet{sun2020conditioned} also prompt by keywords to control for sentiment or topic of the generated sentence. 
In natural language understanding tasks, prompt engineering has been explored in prior works for models like BERT and RoBERTa \citep{liu-etal-roberta, jiang-etal-2020-know, schick2020exploiting}. For example, AutoPrompt \cite{shin2020autoprompt} searches for a sequence of discrete trigger words and concatenates it with each input to elicit sentiment or factual knowledge from a masked LM. 
In contrast with AutoPrompt, our method optimizes continuous prefixes, which are more expressive (\cref{ssec:embed_only}); moreover, we focus on language generation tasks. \looseness=-1


Continuous vectors have been used to steer language models; for example, \citet{subramani2020unconditional} showed that a pretrained LSTM language model can reconstruct arbitrary sentences by optimizing a continuous vector for each sentence, making the vector \emph{input-specific}. In contrast, prefix-tuning optimizes a \emph{task-specific} prefix that applies to all instances of that task. As a result, unlike the previous work whose application is limited to sentence reconstruction, prefix-tuning can be applied to NLG tasks. 
 

\paragraph{Controllable generation.} 
 Controllable generation aims to steer a pretrained language model to match a sentence level attribute (e.g., positive sentiment or topic on sports). Such control can happen at training time: \citet{Keskar2019CTRL} pretrains the language model (CTRL) to condition on metadata such as keywords or URLs. Additionally, the control can happen at decoding time, by weighted decoding \citep[GeDi,][]{KrauseGeDi2020} or iteratively updating the past activations \citep[PPLM,][]{Dathathri2020Plug}. 
 However, there is no straightforward way to apply these controllable generation techniques to enforce fine-grained control over generated contents, as demanded by tasks like table-to-text and summarization.

\section{Problem Statement} 
\begin{figure*}
    \centering
    \includegraphics[width=1.0\textwidth]{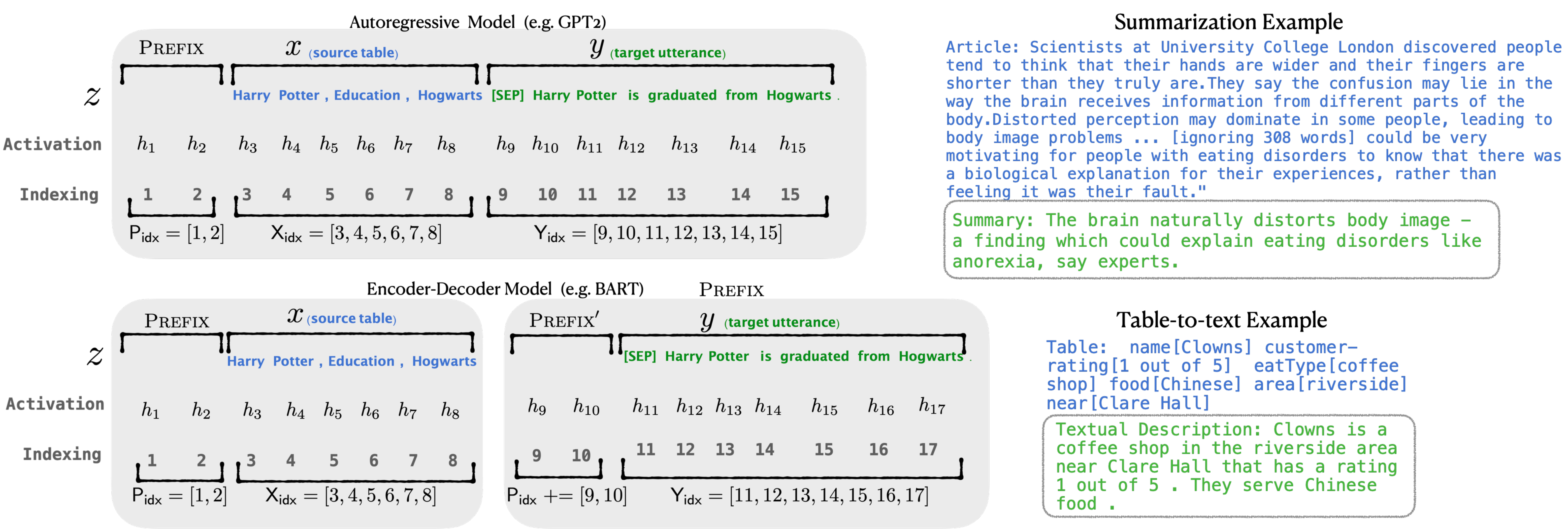}
    \caption{\label{fig:2} An annotated example of prefix-tuning using an autoregressive LM (top) and an encoder-decoder model (bottom). 
    The prefix activations $\forall i \in \pp, h_i$ 
    are drawn from a trainable matrix $\PrefixMatrix_\theta$. The remaining activations are computed by the Transformer.}
    \vspace{-15pt}
\end{figure*}
\label{sec:problem_statement}
Consider a conditional generation task where the input is a context $x$ and the output $y$ is a sequence of tokens. 
We focus on two tasks, shown in \cref{fig:2} (right): In table-to-text, $x$ corresponds to a linearized data table and $y$ is a textual description; in summarization, $x$ is an article and $y$ is a short summary.

\subsection{Autoregressive LM}
Assume we have an autoregressive language model $p_\phi(y \mid x)$ based on the Transformer \cite{Vaswani2017Attn} architecture \citep[e.g., GPT-2;][]{Radford2019gpt2} and parametrized by $\phi$. As shown in \cref{fig:2} (top), let $z = [x;y]$ be the concatenation of $x$ and $y$; let $\xx$ denote the sequence of indices that corresponds to $x$, and $\yy$ denote the same for $y$. 

The activation at time step $i$ is $h_i \in \mathbb R^d$, where $h_i = [h_i^{(1)}; \cdots ; h_i^{(n)}]$ is a concatenation of all activation layers at this time step, and $h_i^{(j)}$ is the activation of the $j$-th Transformer layer at time step $i$.\footnote{$h_i^{(n)}$ is composed of a key-value pair. In GPT-2, the dimension of each key and value is $1024$.} \looseness=-1

The autoregressive Transformer model computes $h_i$ as a function of $z_i$ and the past activations in its left context, as follows: 
\begin{align}
h_i = \LM_\phi(z_i, h_{<i}) \text{,}
\label{eqn:recur}
\end{align}
where the last layer of $h_i$ is used to compute the distribution for the next token: $p_{\phi}(z_{i+1} \mid h_{\leq i}) = \Softmax(W_\phi ~h_{i}^{(n)})$
and $W_\phi$ is a pretrained matrix that map $h_{i}^{(n)}$ to logits over the vocabulary. 

\subsection{Encoder-Decoder Architecture}
We can also use an encoder-decoder architecture \citep[e.g., BART;][]{lewis-etal-2020-bart} to model $p_\phi(y\mid x)$, where $x$ is encoded by the bidirectional encoder, and the decoder predicts $y$ autoregressively (conditioned on the encoded $x$ and its left context). We use the same indexing and activation notation, as shown in \cref{fig:2} (bottom). 
$h_i$ for all $i \in \xx$ is computed by the bidirectional Transformer encoder; 
$h_i$ for all $i \in \yy$ is computed by the autoregressive decoder using the same \cref{eqn:recur}.

\subsection{Method: Fine-tuning}
In the fine-tuning framework, we initialize with the pretrained parameters $\phi$. Here $p_{\phi}$ is a trainable language model distribution and we perform gradient updates on the following log-likelihood objective:

\begin{align}
    \max_{\phi} ~\log p_\phi(y \mid x) = \sum_{i\in \yy}  \log p_\phi(z_i \mid h_{<i}) \text{.}
    \label{eqn:loss}
\end{align}

\section{Prefix-Tuning}

We propose prefix-tuning as an alternative to fine-tuning for conditional generation tasks. We first provide intuition in \cref{ssec:intuition} before defining our method formally in \cref{ssec:formalmethod}.

\subsection{Intuition} 
\label{ssec:intuition}
Based on intuition from prompting, we believe that having a proper context can steer the LM without changing its parameters. For example, if we want the LM to generate a word (e.g., Obama), we can prepend its common collocations as context (e.g., Barack), and the LM will assign much higher probability to the desired word.
Extending this intuition beyond generating a single word or sentence, we want to find a context that steers the LM to solve an NLG task. Intuitively, the context can influence the encoding of $x$ by guiding what to extract from $x$; and can influence the generation of $y$ by steering the next token distribution. However, it's non-obvious whether such a context exists. 
Natural language task instructions (e.g., ``summarize the following table in one sentence'') might guide an expert annotator to solve the task, but fail for most pretrained LMs.\footnote{In our preliminary experiments, GPT-2 and BART fail in this setting; the only exception is GPT-3. } Data-driven optimization over the discrete instructions might help, but discrete optimization is computationally challenging. 

Instead of optimizing over discrete tokens, we can optimize the instruction as continuous word embeddings, whose effects will be propagated upward to all Transformer activation layers and rightward to subsequent tokens. This is strictly more expressive than a discrete prompt which requires matching the embedding of a real word. Meanwhile, this is less expressive than intervening all layers of the activations (\cref{ssec:embed_only}), which avoids long-range dependencies and includes more tunable parameters. Prefix-tuning, therefore, optimizes all layers of the prefix.

\subsection{Method}
\label{ssec:formalmethod}
Prefix-tuning prepends a prefix for an autoregressive LM to obtain $z = [\prefix; x; y]$,  
or prepends prefixes for both encoder and encoder to obtain $z = [\prefix; x; \prefix'; y]$, as shown in \cref{fig:2}. Here, $\pp$ denotes the sequence of prefix indices, and we use $|\pp|$ to denote the length of the prefix. 

We follow the recurrence relation in \cref{eqn:recur}, except that the prefix are \emph{free} parameters. 
Prefix-tuning initializes a trainable matrix $\PrefixMatrix_\theta$ (parametrized by $\theta$) of dimension $|\pp| \times \dim(h_i)$ to store the prefix parameters.
\begin{align}
h_{i} = 
\begin{cases}
\PrefixMatrix_{\theta}[i,:],             & \text{if } i \in \pp \text{,}\\
\LM_\phi(z_i, h_{<i}),    	      & \text{otherwise.}
\end{cases}
\end{align}
The training objective is the same as \cref{eqn:loss}, but the set of trainable parameters changes: the language model parameters $\phi$ are fixed and the prefix parameters $\theta$ are the only trainable parameters. 

Here, $h_i$ (for all $i$) is a function of the trainable $\PrefixMatrix_\theta$. 
When $i \in \pp $, this is clear because $h_i$ copies directly from $\PrefixMatrix_{\theta}$. 
When $i \not \in  \pp $, $h_i$ still depends on $\PrefixMatrix_{\theta}$, because the prefix activations are always in the left context and will therefore affect any activations to its right.

\subsection{Parametrization of $\PrefixMatrix_\theta$ }
\label{ssec:parametrization}
Empirically, directly updating the $\PrefixMatrix_\theta$ parameters leads to unstable optimization and a slight drop in performance.\footnote{We find in preliminary experiments that directly optimizing the prefix is very sensitive to the learning rate and initialization.}
So we reparametrize the matrix $\PrefixMatrix_\theta [i,:]= \MLP_\theta (\PrefixMatrix'_\theta[i,:])$ by a smaller matrix ($\PrefixMatrix'_\theta$) composed with a large feedforward neural network ($\MLP_\theta$). 
Note that  $\PrefixMatrix_\theta$ and $\PrefixMatrix'_\theta$ has the same rows dimension (i.e. the prefix length), but different columns dimension.\footnote{$\PrefixMatrix_\theta$  has a dimension of $|\pp| \times \dim(h_i)$ while $\PrefixMatrix_\theta$ has a dimension  of  $|\pp| \times k$, where we choose $k=512$ for table-to-text and $800$ for summarization. $\MLP_\theta$ maps from dimension $k$ to $\dim(h_i)$} 
Once training is complete, these reparametrization parameters can be dropped, and only the prefix ($\PrefixMatrix_\theta$) needs to be saved.  

\section{Experimental Setup}

\subsection{Datasets and Metrics}
\label{ssec:dataset}
We evaluate on three standard neural generation datasets for the table-to-text task: E2E \cite{e2edata}, WebNLG \cite{webnlg-2017}, and DART \cite{radev2020dart}. The datasets are ordered by increasing complexity and size. E2E only has $1$ domain (i.e. restaurant reviews); WebNLG has $14$ domains, and DART is open-domain, using open-domain tables from Wikipedia.

The E2E dataset contains approximately 50K examples with 8 distinct fields; it contains multiple test references for one source table, and the average output length is $22.9$.  We use the official evaluation script, which reports  BLEU \cite{bleu}, NIST \cite{belz-reiter-nist}, METEOR \cite{Lavie-meteor}, ROUGE-L \cite{lin-2004-rouge}, and CIDEr \cite{cider}.

The WebNLG \cite{webnlg-2017} dataset consists of 22K
examples, and the input $x$ is a sequence of (subject, property, object) triples. The average output length is $22.5$. In the training and validation splits, the input describes entities from $9$ distinct DBpedia categories (e.g., Monument). 
The test split consists of two parts: the first half contains DB categories seen in training data, and the second half contains $5$ unseen categories.
These unseen categories are used to evaluate extrapolation. We use the official evaluation script, which reports BLEU, METEOR and TER \cite{ter}.  

DART \cite{radev2020dart} is an open domain table-to-text dataset, with similar input format (entity-relation-entity triples) as WebNLG. The average output length is $21.6$. It consists of 82K
examples from WikiSQL, WikiTableQuestions, E2E, and WebNLG and applies some manual or automated conversion. We use the official evaluation script and report BLEU, METEOR, TER, MoverScore \cite{zhao-etal-2019-moverscore}, BERTScore \cite{bert-score} and BLEURT \cite{sellam-etal-2020-bleurt}. 

For the summarization task, we use the XSUM \cite{xsum-emnlp} dataset, which is an abstractive summarization dataset on news articles. There are 225K examples. The average length of the articles is 431 words and the average length of the summaries is 23.3. We report ROUGE-1, ROUGE-2 and ROUGE-L. 

\subsection{Methods}
For table-to-text generation, we compare prefix-tuning with three other methods: fine-tuning (\FT), fine-tuning only the top 2 layers (\FTtop), and adapter-tuning (\adapter).\footnote{Same implementation as \citet{lin-etal-2020-exploring}.} We also report the current state-of-the-art results on these datasets: On E2E, \citet{shen-etal-2019-pragmatically} uses a pragmatically informed model without pretraining. On WebNLG, \citet{kale2020texttotext} fine-tunes T5-large. On DART, no official models trained on this dataset version are released.\footnote{The official benchmark model is trained on v.1.0.0 while the release dataset is v1.1.1.} For summarization, we compare against fine-tuning BART \cite{lewis-etal-2020-bart}.

\subsection{Architectures and Hyperparameters}
For table-to-text, we use GPT-2$_{\textsc{MEDIUM}}$ and GPT-2$_{\textsc{LARGE}}$; the source tables are linearized.\footnote{In comparison with natural language utterances, the linearized table is in an unnatural format, which might be challenging for pretrained LMs.} 
For summarization, we use BART$_{\textsc{LARGE}}$,\footnote{We didn't include GPT-2 results for summarization because in our preliminary experiment, fine-tuning GPT-2 significantly underperforms fine-tuning BART on XSUM.} and the source articles are truncated to $512$ BPE tokens. 

Our implementation is based on the Hugging Face Transformer models \cite{wolf-etal-2020-transformers}. At training time, we use the AdamW optimizer \cite{loshchilov2018decoupled} and a linear learning rate scheduler, as suggested by the Hugging Face default setup.  The hyperparameters we tune include the number of epochs, batch size, learning rate, and prefix length. Hyperparameter details are in the appendix. A default setting trains for $10$ epochs, using a batch size of $5$, a learning rate of $5 \cdot 10^{-5}$ and a prefix length of $10$. The table-to-text models are trained on TITAN Xp or GeForce GTX TITAN X machines. 
Prefix-tuning takes $0.2$ hours per epochs to train on 22K examples , whereas fine-tuning takes around $0.3$ hours. 
The summarization models are trained on Tesla V100 machines, taking $1.25$h per epoch on the XSUM dataset. 

At decoding time, for the three table-to-text datasets, we use beam search with a beam size of $5$. For summarization, we use a beam size of $6$ and length normalization of $0.8$. 
Decoding takes $1.2$ seconds per sentence (without batching) for table-to-text, and $2.6$ seconds per batch (using a batch size of 10) for summarization. 
\section{Main Results}

\label{sec:experiment}
\begin{table*}[]
    \centering
    \resizebox{2.1\columnwidth}{!}{
    \addtolength{\tabcolsep}{-2pt} 
    \begin{tabular}{lccccc|ccccccccc|cccccc}
\toprule 
& \multicolumn{5}{c}{E2E} & \multicolumn{9}{|c}{WebNLG} & \multicolumn{6}{|c}{DART} \\

         & BLEU  & NIST & MET & R-L & CIDEr  & \multicolumn{3}{c}{BLEU}  & \multicolumn{3}{c}{MET} & \multicolumn{3}{c|}{TER $\downarrow$}  & BLEU   & MET   & TER $\downarrow$  & Mover    & BERT   & BLEURT \\
& \multicolumn{5}{c|}{ } & S & U & A & S & U & A & S & U & A \\
\midrule
\midrule
& \multicolumn{20}{c}{ GPT-2$_{\textsc{MEDIUM}}$} \\
\FT       & 68.2 & 8.62 & \textbf{46.2}  & 71.0  & 2.47 & 
\textbf{64.2}  & 27.7 & 46.5 & \textbf{0.45} & 0.30 & 0.38 & \textbf{0.33} & 0.76 & 0.53 & 
46.2 & \textbf{0.39} & \textbf{0.46} & \textbf{0.50} & \textbf{0.94} & \textbf{0.39} \\ 
\FTtop      & 68.1 & 8.59 & 46.0 & 70.8  & 2.41 & 
53.6 & 18.9    & 36.0 & 0.38 & 0.23 & 0.31 & 0.49 & 0.99 & 0.72  & 
41.0 & 0.34 & 0.56 & 0.43 & 0.93 & 0.21 \\ %
\adapter (3\%)   & 68.9 & 8.71 & 46.1  & 71.3  & 2.47 & 
60.4 & \textbf{48.3} & 54.9 & 0.43 & \textbf{0.38} & \textbf{0.41} & 0.35 & \textbf{0.45} & \textbf{0.39} & 
45.2 & 0.38 & \textbf{0.46} & \textbf{0.50} & \textbf{0.94} & \textbf{0.39} \\
\adapter (0.1\%) & 66.3 & 8.41 & 45.0  & 69.8  & 2.40 & 
54.5 & 45.1 & 50.2 & 0.39 & 0.36 & 0.38 & 0.40 & 0.46 & 0.43 & 
42.4 & 0.36 & 0.48 & 0.47 & \textbf{0.94} & 0.33 \\ 
\prefix  (0.1\%) & \textbf{69.7} & \textbf{8.81} & 46.1  & \textbf{71.4}  & \textbf{2.49} & 62.9 & 45.6 & \textbf{55.1} & 0.44 & \textbf{0.38} & \textbf{0.41} & 0.35 & 0.49 & 0.41 & 
\textbf{46.4} & 0.38 & \textbf{0.46} & \textbf{0.50} & \textbf{0.94} & \textbf{0.39} \\ 
\midrule
& \multicolumn{20}{c}{ GPT-2$_{\textsc{LARGE}}$} \\
\FT       & 68.5 & 8.78 & 46.0 & 69.9 & 2.45 & \textbf{65.3} & 43.1 & 55.5 & \textbf{0.46} & 0.38 & \textbf{0.42} & \textbf{0.33} & 0.53 & 0.42  
& \textbf{47.0} & \textbf{0.39} & 0.46 & \textbf{0.51} & \textbf{0.94} & \textbf{0.40} \\ 
Prefix          & \textbf{70.3}  & \textbf{8.85}  & \textbf{46.2}  & \textbf{71.7}  & \textbf{2.47} & 63.4 & \textbf{47.7} & \textbf{56.3} & 0.45 & \textbf{0.39} & \textbf{0.42} & 0.34 & \textbf{0.48} & \textbf{0.40} 
& 46.7 & \textbf{0.39} & \textbf{0.45} & \textbf{0.51} & \textbf{0.94} & \textbf{0.40} \\
\midrule
SOTA            & 68.6 & 8.70 & 45.3 &70.8 & 2.37  &  63.9 & 52.8 & 57.1 & 0.46 & 0.41 & 0.44 & - & - & - & - & - & - & - & - & - \\

\bottomrule
\end{tabular}
}
\caption{\label{tab:table-to-text} Metrics (higher is better, except for TER) for table-to-text generation on E2E (left), WebNLG (middle) and DART (right). With only $0.1\%$ parameters, Prefix-tuning outperforms other lightweight baselines and achieves a comparable performance  with fine-tuning. The best score is boldfaced for both GPT-2$_{\textsc{MEDIUM}}$ and GPT-2$_{\textsc{LARGE}}$. }
\end{table*}

\subsection{Table-to-text Generation}
\label{ssec:table-to-text}

We find that adding only 0.1\% task-specific parameters,\footnote{250K for E2E, 250K for WebNLG, and 500K for DART vs. 345M GPT-2 parameters.}
prefix-tuning is effective in table-to-text generation, outperforming other lightweight baselines ($\adapter$ and $\FTtop$) and achieving a comparable performance with fine-tuning. This trend is true across all three datasets: E2E, WebNLG,\footnote{The S,U,A columns in WebNLG represents SEEN, UNSEEN, and ALL respectively; \underline{S}EEN categories appear at training time; \underline{U}NSEEN categories only appears at test time; and \underline{A}LL is the combination of the two.} and DART.

For a fair comparison, we match the number of parameters for prefix-tuning and adapter-tuning to be 0.1\%. \cref{tab:table-to-text} shows that prefix-tuning is significantly better than $\adapter$ (0.1\%), attaining $4.1$ BLEU improvement per dataset on average. 
Even when we compare with fine-tuning (100\%) and adapter-tuning (3.0\%),
which update significantly more parameters than prefix-tuning,
prefix-tuning still achieves results comparable or better than those two systems. 
This demonstrates that prefix-tuning is more Pareto efficient than adapter-tuning, significantly reducing parameters while improving generation quality. 

Additionally, attaining good performance on DART suggests that prefix-tuning can generalize to tables with diverse domains and a large pool of relations. We will delve deeper into extrapolation performance (i.e. generalization to unseen categories or topics) in \cref{sec:extrapolation}. 

Overall, prefix-tuning is an effective and space-efficient method to adapt GPT-2 to table-to-text generation. The learned prefix is expressive enough to steer GPT-2 in order to correctly extract contents from an unnatural format 
and generate a textual description. 
Prefix-tuning also scales well from GPT-2$_{\textsc{MEDIUM}}$ to GPT-2$_{\textsc{LARGE}}$, suggesting it has the potential to scale to even larger models with a similar architecture, like GPT-3. 

\begin{figure*}[hbt!]
\begin{minipage}{1.0\columnwidth}
\resizebox{1.05\columnwidth}{!}{
    \begin{tabular}{lp{12cm}}
    \toprule
        Source & name : The Eagle $|$ type : coffee shop $|$ food : Chinese $|$ price : cheap $|$ customer rating : average $|$  area : riverside $|$  family friendly : no $|$ near : Burger King \\
        \midrule
        Prefix (50) & The Eagle is a cheap Chinese coffee shop located near Burger King. \\
        Prefix (100)   &  The Eagle is a cheap coffee shop located in the riverside near Burger King. It has average customer ratings.\\
        Prefix (200)  & The Eagle is a cheap Chinese coffee shop located in the riverside area near Burger King. It has average customer ratings.\\
        Prefix (500)  & The Eagle is a coffee shop that serves Chinese food. It is located in the riverside area near Burger King. It has an average customer rating and is not family friendly.\\
        \midrule
        FT (50) & The Eagle coffee shop is located in the riverside area near Burger King.\\
        FT (100) &  The Eagle is a cheap coffee shop near Burger King in the riverside area. It has a low customer rating and is not family friendly. \\
        FT (200) &  The Eagle is a cheap Chinese coffee shop with a low customer rating. It is located near Burger King in the riverside area. \\
        FT (500) &  The Eagle is a cheap Chinese coffee shop with average customer ratings. It is located in the riverside area near Burger King.
\\  
        \bottomrule
    \end{tabular}}
\end{minipage}
\begin{minipage}{1.1\columnwidth}
 \centering
    \includegraphics[page=1,width=0.45\textwidth]{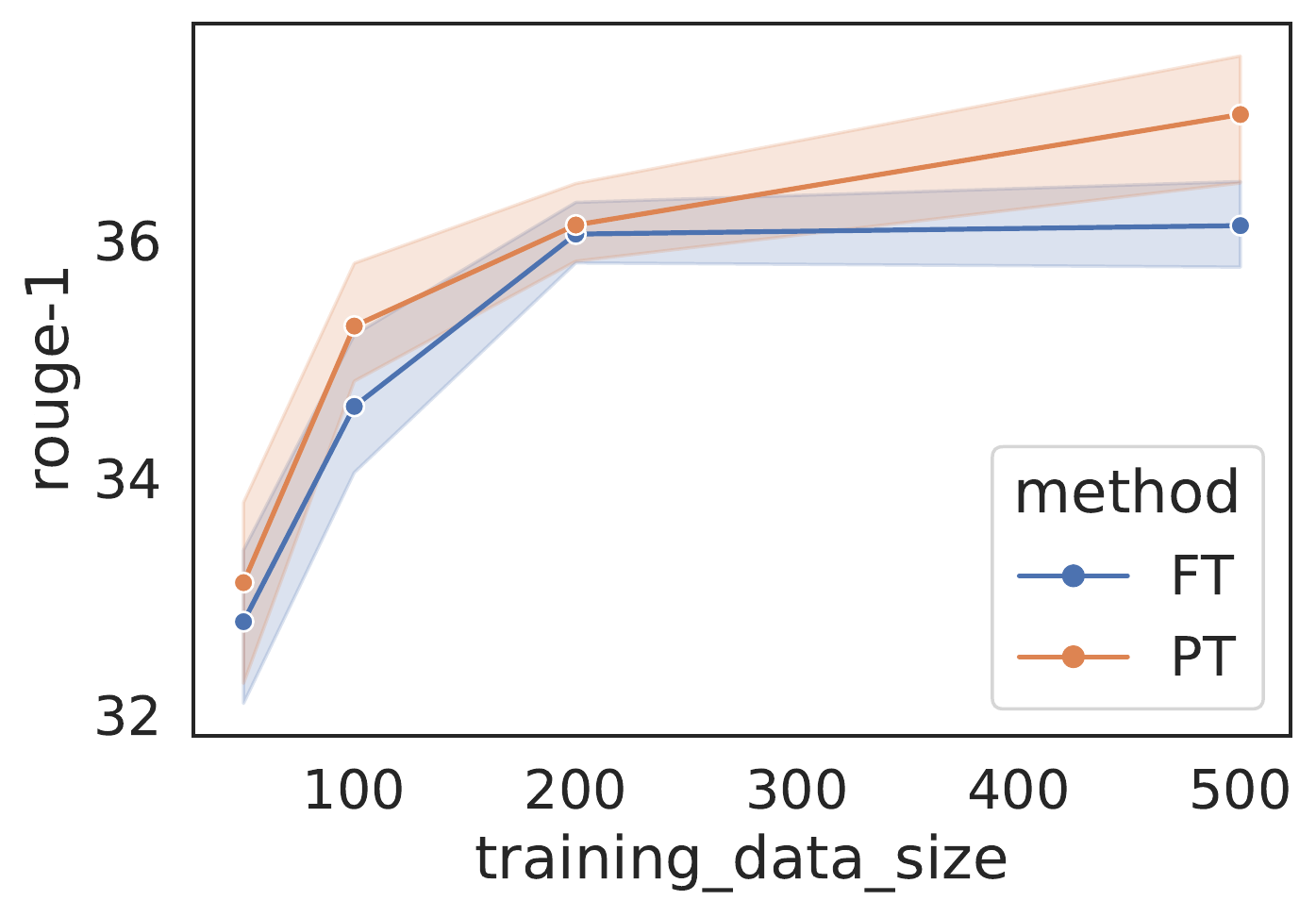} \hspace{-20pt}
    \qquad
    \includegraphics[page=2,width=0.45\textwidth]{xsum_lowdata.pdf} \hspace{-20pt}
    \\ 
\hspace{-25pt}
\centering
    \includegraphics[page=1,width=0.45\textwidth]{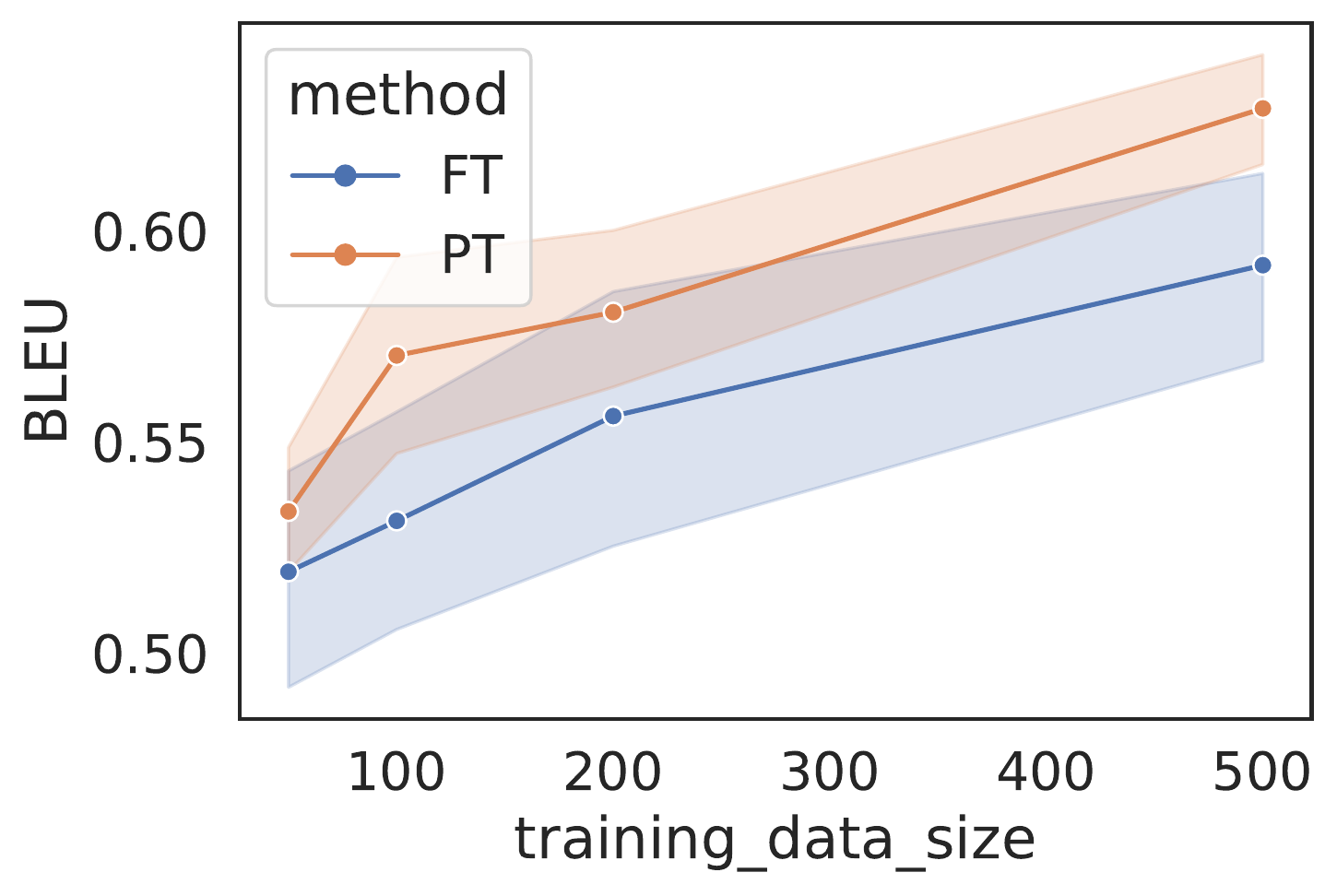} \hspace{-20pt}
    \qquad
    \includegraphics[page=3,width=0.45\textwidth]{e2e_lowdata.pdf} \hspace{-20pt}
\end{minipage}
\caption{\label{tab:lowdata} (Left) qualitative examples in lowdata settings. (Right) prefix-tuning (orange) outperforms fine-tuning (blue) in low-data regimes in addition to requiring many fewer parameters. The top two plots correspond to summarization, measured by ROUGE-1 and ROUGE-2. The bottom two plots correspond to table-to-text, measured by BLEU and ROUGE-L. The x-axis is the training size and the y-axis is the evaluation metric (higher is better).}
\vspace{-12pt}
\end{figure*}

\subsection{Summarization}
\label{ssec:summarization}

\begin{table}[hbt!]
\resizebox{\columnwidth}{!}{
\begin{tabular}{lllll}
\toprule
                 & R-1 $\uparrow$  & R-2 $\uparrow$  & R-L $\uparrow$  \\ \hline
\FT \cite{lewis-etal-2020-bart}   & 45.14   & 22.27 & 37.25 \\ 
\prefix (2\%)  & 43.80   & 20.93 & 36.05 \\ 
\prefix (0.1\%)  & 42.92   & 20.03 & 35.05 \\ 
\bottomrule
\end{tabular}}
\caption{\label{tab:xsum} Metrics for summarization on XSUM. Prefix-tuning slightly underperforms fine-tuning.}
\end{table}
As shown in \cref{tab:xsum}, with 2\% parameters, prefix-tuning obtains slightly lower performance than fine-tuning (36.05 vs.\@ 37.25 in ROUGE-L). 
With only 0.1\% parameters, prefix-tuning underperforms full fine-tuning (35.05 vs. 37.25). 
There are several differences between XSUM and the three table-to-text datasets which could account for why prefix-tuning has comparative advantage in table-to-text: (1) XSUM contains 4x more examples than the three table-to-text datasets on average; (2) the input articles are 17x longer than the linearized table input of table-to-text datasets on average; (3) summarization might be more complex than table-to-text because it requires reading comprehension and identifying key contents from an article. 

\subsection{Low-data Setting}
\label{sec:lowdata}

Based on the results from table-to-text (\cref{ssec:table-to-text}) and summarization (\cref{ssec:summarization}), we observe that 
prefix-tuning has a comparative advantage when the number of training examples is smaller.
To construct low-data settings, we subsample the full dataset (E2E for table-to-text and XSUM for summarization) to obtain small datasets of size $\{ 50, 100, 200, 500 \}$. For each size, we sample $5$ different datasets and average over $2$ training random seeds. Thus, we average over $10$ models to get an estimate for each low-data setting.\footnote{We also sample a dev split (with dev size = 30\% $\times$ training size ) for each training set. We use the dev split to choose hyperparameters and do early stopping.}

\cref{tab:lowdata} (right) shows that prefix-tuning outperforms fine-tuning in low-data regimes by $2.9$ BLEU on average,
in addition to requiring many fewer parameters, but the gap narrows as the dataset size increases. 

Qualitatively, \Cref{tab:lowdata} (left) shows $8$ examples generated by both prefix-tuning and fine-tuning models trained on different data levels. While both methods tend to undergenerate (missing table contents) in low data regimes, prefix-tuning tends to be more faithful than fine-tuning.  For example, fine-tuning (100, 200)\footnote{The number in the parenthesis refers to the training size.} falsely claims a low customer rating while the true rating is average, whereas prefix-tuning (100, 200) generates a description that is faithful to the table. 

\subsection{Extrapolation}
\label{sec:extrapolation}

We now investigate extrapolation performance to unseen topics for both table-to-text and summarization. 
In order to construct an extrapolation setting, we split the existing datasets so that training and test cover different topics.  
For table-to-text, the WebNLG dataset is labeled with table topics. There are 
$9$ categories that appear in training and dev, denoted as SEEN and
$5$ categories that only appear at test time, denoted as UNSEEN. So we evaluate extrapolation by training on the SEEN categories and testing on the UNSEEN categories.
For summarization, we construct two extrapolation data splits\footnote{XSUM dataset is drawn from BBC news, and we identify the topic of each article based on their URLs. Since ``news'' and ``sports'' are the two domains with the most articles, we create our first train/test split. Additionally, ``news'' has subdomains such as ``UK'', ``world'', and ``technology''. Consequently, we create a second data split, using the top 3 news subdomains as training data and the rest as test data.}: In \texttt{news-to-sports}, we train on news articles, and test on sports articles. In \texttt{within-news}, we train on $\{$world, UK, business$\}$ news, and test on the remaining news categories (e.g., health, technology).

\begin{table}  
\noindent \resizebox{\columnwidth}{!}{
\begin{tabular}{llll|lll}
\toprule
          & \multicolumn{3}{c}{\texttt{news-to-sports}}  & \multicolumn{3}{|c}{\texttt{within-news}} \\
          & R-1 $\uparrow$  & R-2 $\uparrow$  & R-L $\uparrow$  & R-1 $\uparrow$  & R-2 $\uparrow$  & R-L $\uparrow$ \\ \hline
\FT & 38.15           & 15.51           & 30.26  & 39.20 & 16.35 & 31.15\\ 
\prefix    & 39.23           & 16.74           & 31.51  & 39.41 & 16.87 & 31.47\\ 
\bottomrule
\end{tabular}} \vspace{1pt}
\caption{\label{tab:xsum-extra} Extrapolation performance on XSUM. Prefix-tuning outperforms fine-tuning on both news-to-sports and within-news splits.
}
\end{table}

On both table-to-text and summarization, prefix-tuning has better extrapolation than fine-tuning under all metrics, as shown in \cref{tab:xsum-extra} and the `U' columns of \cref{tab:table-to-text} (middle). 

We also find that adapter-tuning achieves good extrapolation performance, comparable with prefix-tuning, as shown in \cref{tab:table-to-text}. This shared trend suggests that preserving LM parameters indeed has a positive impact on extrapolation. However, the reason for such gains is an open question and we will discuss further in \cref{sec:discussion}. 

\section{Intrinsic Evaluation}
\label{sec:intrinsic}
We compare different variants of prefix-tuning. 
\cref{ssec:prefix_length} studies the impact of the prefix length.
\cref{ssec:embed_only} studies tuning only the embedding layer, which is more akin to tuning a discrete prompt. 
\cref{ssec:infixing} compares prefixing and infixing, which inserts trainable activations between $x$ and $y$.  
\cref{ssec:init} studies the impact of various prefix initialization strategies.

\subsection{Prefix Length}
\label{ssec:prefix_length}

\begin{figure}
\hspace{-15pt}
    \centering
    \includegraphics[page=1,width=0.245\textwidth]{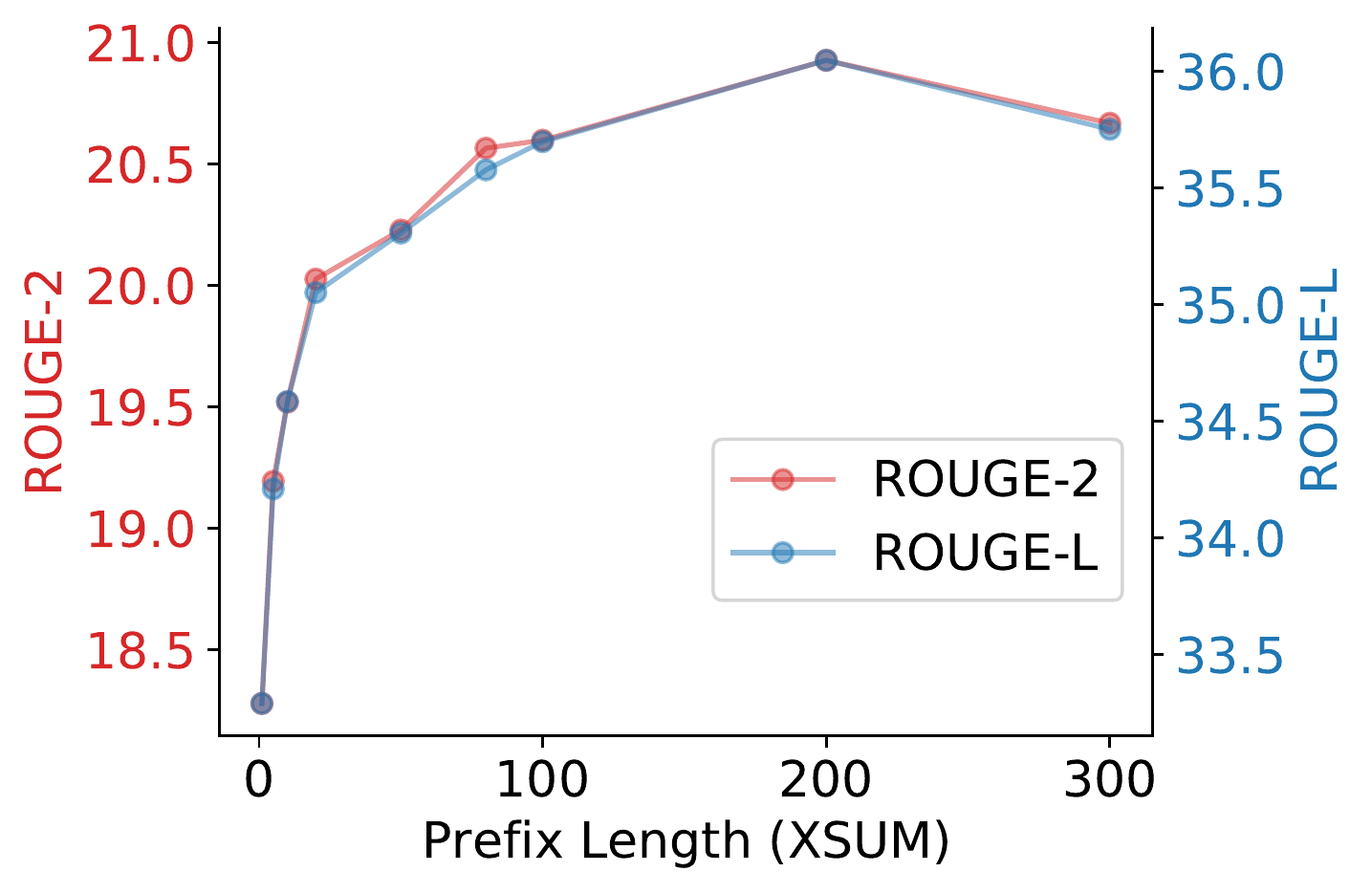} \hspace{-20pt}
    \qquad
    \includegraphics[page=2,width=0.245\textwidth]{prefixlen/prefixlen.pdf} \hspace{-20pt}
\hspace{-15pt}
    \caption{\label{fig:prefixlen} Prefix length vs.\@ performance on summerization (left) and table-to-text (right). Performance increases as the prefix length increases up to a threshold (200 for summarization and 10 for table-to-text) and then a slight performance drop occurs. 
    Each plot reports two metrics (on two vertical axes).}
    
\end{figure}
A longer prefix means more trainable parameters, and therefore more expressive power. 
\cref{fig:prefixlen} shows that performance increases as the prefix length increases up to a threshold ($200$ for summarization, $10$ for table-to-text) and then a slight performance drop occurs.\footnote{Prefixes longer than the threshold lead to lower training loss, but slightly worse test performance, suggesting that they tend to overfit the training data.}

Empirically, longer prefixes have a negligible impact on inference speed, because attention computation over the entire prefix is parallellized on GPUs. 

\subsection{Full \@vs Embedding-only}
\label{ssec:embed_only}

\begin{table}[]
\centering
\small
\begin{tabular}{lccccc}
\toprule
& \multicolumn{5}{c}{E2E}\\
         & BLEU  & NIST & MET & ROUGE & CIDEr \\
\midrule
\midrule
\prefix &  69.7 & 8.81 & 46.1  & 71.4  & 2.49  \\
\midrule
& \multicolumn{5}{c}{Embedding-only: \Emb-\{PrefixLength\}}\\
\Emb-1 & 48.1 & 3.33 & 32.1 & 60.2 & 1.10 \\ 
\Emb-10 & 62.2 & 6.70 & 38.6& 66.4 & 1.75 \\ 
\Emb-20 & 61.9 & 7.11 & 39.3 & 65.6 & 1.85 \\ 
\midrule
& \multicolumn{5}{c}{Infix-tuning: \infix-\{PrefixLength\}}\\
\infix-1 &  67.9 & 8.63 & 45.8 & 69.4 & 2.42 \\
\infix-10 &  67.2 & 8.48 & 45.8 & 69.9 & 2.40 \\
\infix-20 & 66.7  & 8.47& 45.8 & 70.0 & 2.42\\
\bottomrule
\end{tabular}
\caption{\label{tab:ablation} Intrinsic evaluation of Embedding-only (\cref{ssec:embed_only}) and Infixing (\cref{ssec:infixing}). Both Embedding-only ablation and Infix-tuning underperforms full prefix-tuning.}
\end{table}
Recall in \cref{ssec:intuition}, we discuss the option of optimizing the continuous embeddings of the ``virtual tokens.'' We instantiate that idea and call it embedding-only ablation. The word embeddings are free parameters, and the upper activation layers are computed by the Transformer. \cref{tab:ablation} (top) shows that the performance drops significantly, suggesting that tuning only the embedding layer is not sufficiently expressive. 

The embedding-only ablation upper bounds the performance of discrete prompt optimization \cite{shin2020autoprompt}, because discrete prompt restricts the embedding layer to exactly match the embedding of a real word. Consequently, we have this chain of increasing expressive power: discrete prompting $<$ embedding-only ablation $<$ prefix-tuning. 

\subsection{Prefixing \@vs Infixing}
\label{ssec:infixing}
We also investigate how the trainable activations' position in the sequence affects performance. In prefix-tuning, we place them at the beginning $[\prefix ; x ; y]$. We can also place the trainable activations between $x$ and $y$ (i.e. $[x;\infix;y]$) and call this infix-tuning. \cref{tab:ablation} (bottom) shows that infix-tuning slightly underperforms prefix-tuning. We believe this is because prefix-tuning can affect the activations of $x$ and $y$ whereas infix-tuning can only influence the activations of $y$.

\subsection{Initialization}
\label{ssec:init}

\begin{figure}
 \includegraphics[page=1,width=0.5\textwidth]{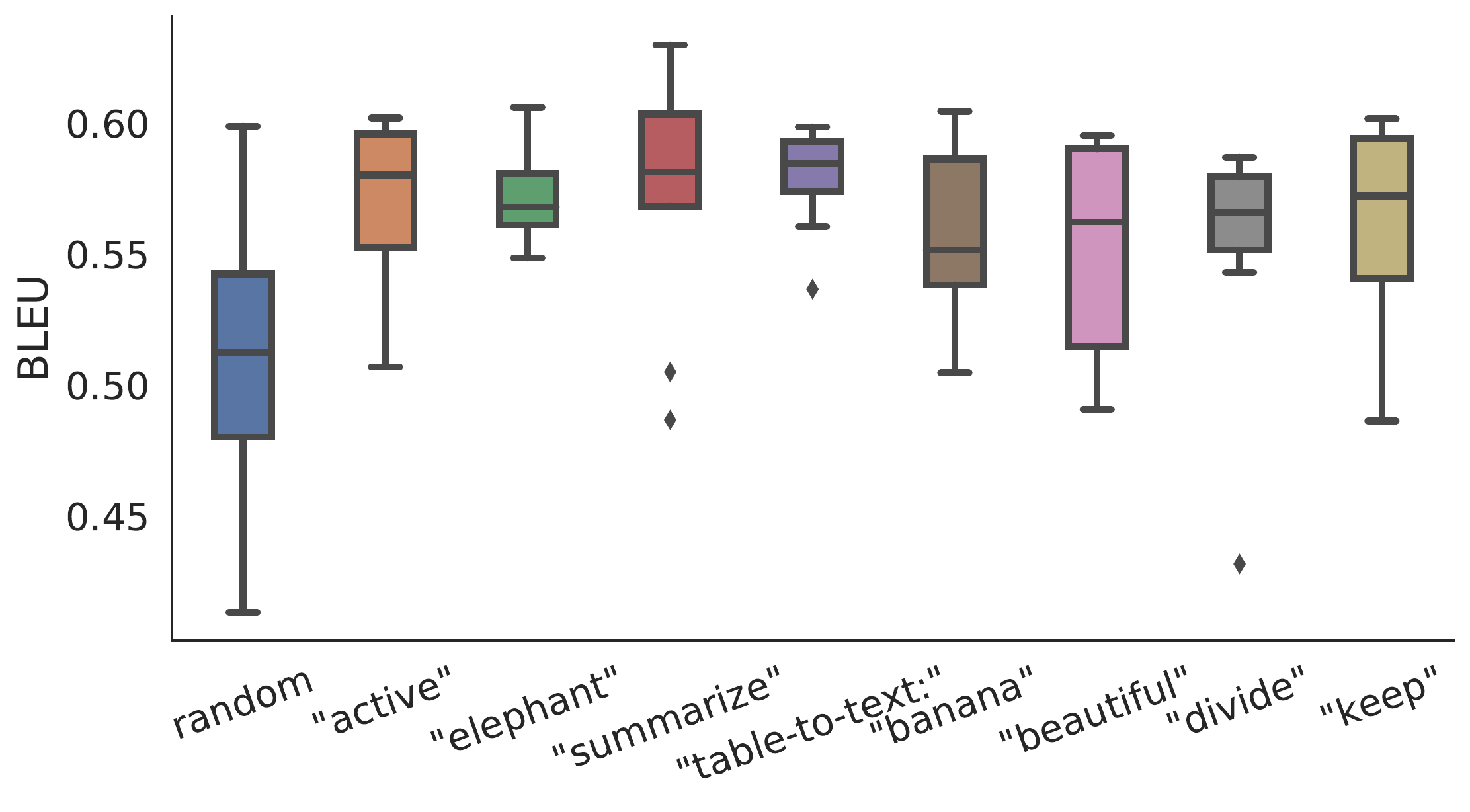}
 \caption{\label{fig:init} Initializing the prefix with activations of real words significantly outperforms random initialization, in low-data settings.}
\end{figure}
We find that how the prefix is initialized has a large impact in low-data settings. Random initialization leads to low performance with high variance. Initializing the prefix with activations of real words significantly improves generation, as shown in \cref{fig:init}. In particular, initializing with task relevant words such as ``summarization'' and ``table-to-text'' obtains slightly better performance than task irrelevant words such as  ``elephant'' and ``divide'', but using real words is still better than random. \looseness=-1

Since we initialize the prefix with activations of real words computed by the LM, this initialization strategy is concordant with preserving the pretrained LM as much as possible. 


\section{Discussion}
\label{sec:discussion}
In this section, we will discuss several favorable properties of prefix-tuning and some open problems. 

\subsection{Personalization} 

As we note in \cref{sec:intro}, prefix-tuning is advantageous when there are a large number of tasks that needs to be trained independently. 
One practical setting is user privacy \cite{Shokri-Shmatikov-2015, McMahan-et-al-2016}. In order to preserve user privacy, each user's data needs to be separated and a personalized model needs to be trained independently for each user. Consequently, each user can be regarded as an independent task. 
If there are millions of users, prefix-tuning can scale to this setting and maintain modularity, enabling flexible addition or deletion of users by adding or deleting their prefixes without cross-contamination. 

\subsection{Batching Across Users}
Under the same personalization setting, prefix-tuning allows batching different users' queries even though they are backed by different prefixes. 
When multiple users query a cloud GPU device with their inputs, it is computationally efficient to put these users in the same batch. 
Prefix-tuning keeps the shared LM intact; consequently, batching requires a simple step of prepending the personalized prefix to user input, and all the remaining computation is unchanged. 
In contrast, we can't batch across different users in adapter-tuning, which has personalized adapters between shared Transformer layers.   


\subsection{Inductive Bias of Prefix-tuning}
Recall that fine-tuning updates all pretrained parameters, whereas prefix-tuning and adapter-tuning preserve them. Since the language models are pretrained on general purpose corpus, preserving the LM parameters might help generalization to domains unseen during training. In concordance with this intuition, we observe that both prefix-tuning and adapter-tuning have significant performance gain in extrapolation settings (\cref{sec:extrapolation}); however, the reason for such gain is an open question. 

While prefix-tuning and adapter-tuning both freeze the pretrained parameters, they tune different sets of parameters to affect the activation layers of the Transformer. Recall that prefix-tuning keeps the LM intact and uses the prefix and the pretrained attention blocks to affect the subsequent activations; adapter-tuning inserts trainable modules between LM layers, which directly add residual vectors to the activations. 
Moreover, we observe that prefix-tuning requires vastly fewer parameters compared to adapter-tuning while maintaining comparable performance. We think this gain in parameter efficiency is because prefix-tuning keeps the pretrained LM intact as much as possible, and therefore exploits the LM more than adapter-tuning.  



Concurrent work by \citet{aghajanyan2020intrinsic} uses intrinsic dimension to show that there exists a low dimension reparameterization that is as effective for fine-tuning as the full parameter space. 
This explains why good accuracy on downstream task can be obtained by updating only a small number of parameters. Our work echoes the finding by showing that good generation performance can be attained by updating a very small prefix.


\section{Conclusion}

We have proposed prefix-tuning, a lightweight alternative to fine-tuning that prepends a trainable continuous prefix for NLG tasks. 
We discover that despite learning 1000x fewer parameters than fine-tuning, prefix-tuning can maintain a comparable performance in a full data setting and outperforms fine-tuning in both low-data and extrapolation settings.



\bibliography{acl2020}
\bibliographystyle{acl_natbib}

\clearpage
\newpage
\appendix

\section{Supplementary Material}
\label{sec:appendix}

\subsection{Hyperparameters} 
\label{ssub:hyperparameters}
In \cref{app:hyper}, we report the hyperparameters used to train the models documented in the experiment section. 

\begin{table}
\centering
\resizebox{\columnwidth}{!}{
\begin{tabular}{lccccc}
\toprule
         & learning rate  & \# epoch & batch size & prefix length \\
\midrule
\midrule
Prefix:\\
E2E & 8e-05  & 5 & 10 & 5  \\
WebNLG & 5e-05 & 5 & 5 & 5 \\ 
DART &  5e-05 & 10 & 5 & 10 \\ 
XSUM &  5e-05 & 30 & 14 & 100 \\ 
\midrule
Adapter:\\
E2E (3\%)     & 5e-05   & 5   & 5 & -  \\
E2E (0.1\%)   & 8e-05   & 10  & 5 \\ 
WebNLG (3\%)  & 5e-05   & 5   & 5 & - \\ 
WebNLG (0.1\%)  & 5e-05   & 10   & 5 & - \\ 
DART (3\%)    & 5e-05   & 5  & 5 & -\\ 
DART (0.1\%)    & 8e-05   & 5  & 5 & -\\ 
\midrule
Fine-tune:\\
E2E & 5e-05  & 5 & 10 & -  \\
WebNLG & 1e-05 & 10 & 6 & - \\ 
DART &  1e-05 & 10 & 6 & - \\ 
\midrule
FT-top2:\\
E2E & 5e-05  & 5 & 10 & -  \\
WebNLG & 5e-05 & 10 & 9 & - \\ 
DART &  5e-05 & 5 & 5 & - \\ 

\end{tabular}}
\caption{\label{app:hyper} Hyperparameter settings for our method and baseline methods. }
\end{table}


\subsection{Additional Results for Low-data Settings}
\begin{figure*}[]
  \centering
    \includegraphics[page=1,width=0.30\textwidth]{xsum_lowdata.pdf} \hspace{-20pt}
    \qquad
    \includegraphics[page=2,width=0.30\textwidth]{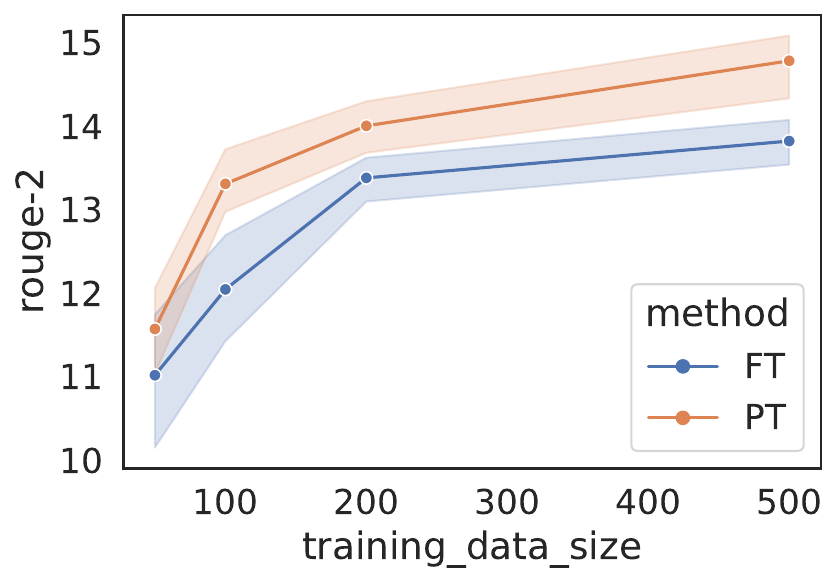} \hspace{-20pt}
    \qquad
    \includegraphics[page=3,width=0.30\textwidth]{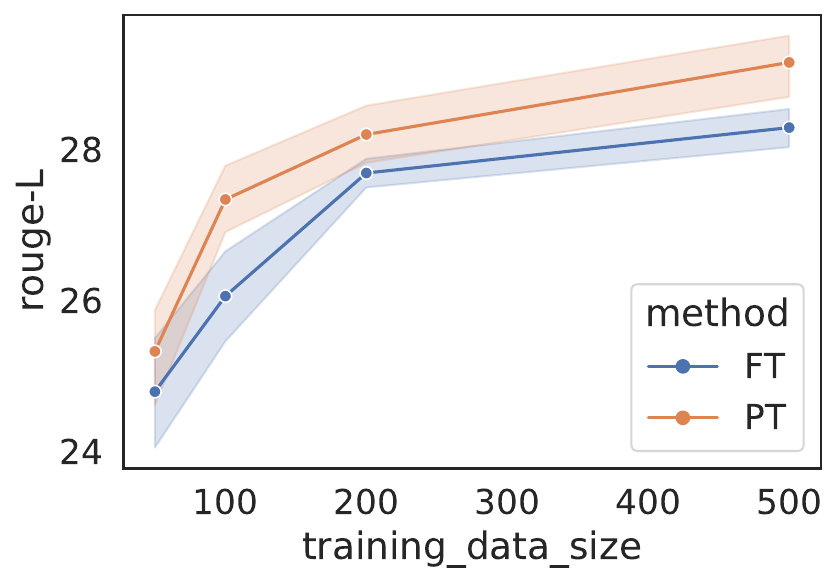} \hspace{-20pt}
    \\ 
\hspace{-25pt}
\centering
    \includegraphics[page=2,width=0.30\textwidth]{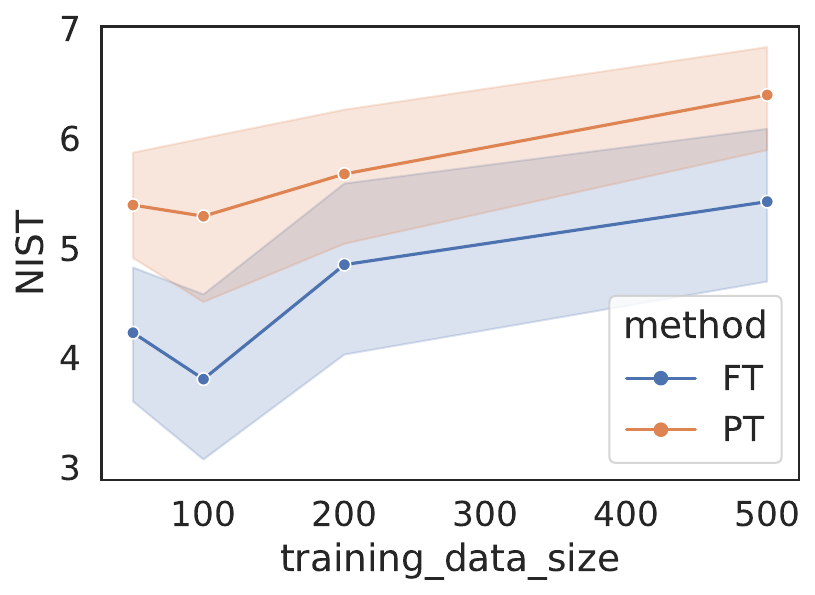} \hspace{-20pt}
    \qquad
    \includegraphics[page=4,width=0.30\textwidth]{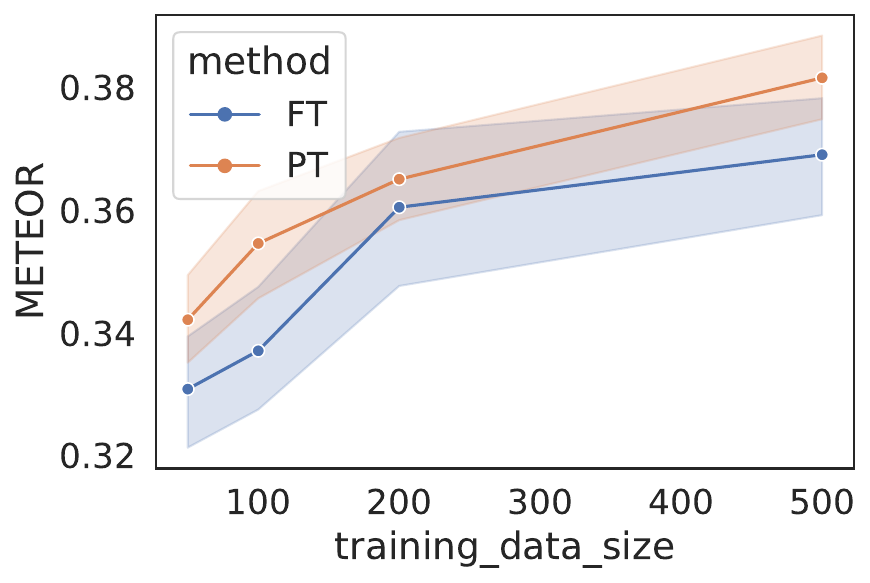} \hspace{-20pt}
    \qquad
    \includegraphics[page=5,width=0.30\textwidth]{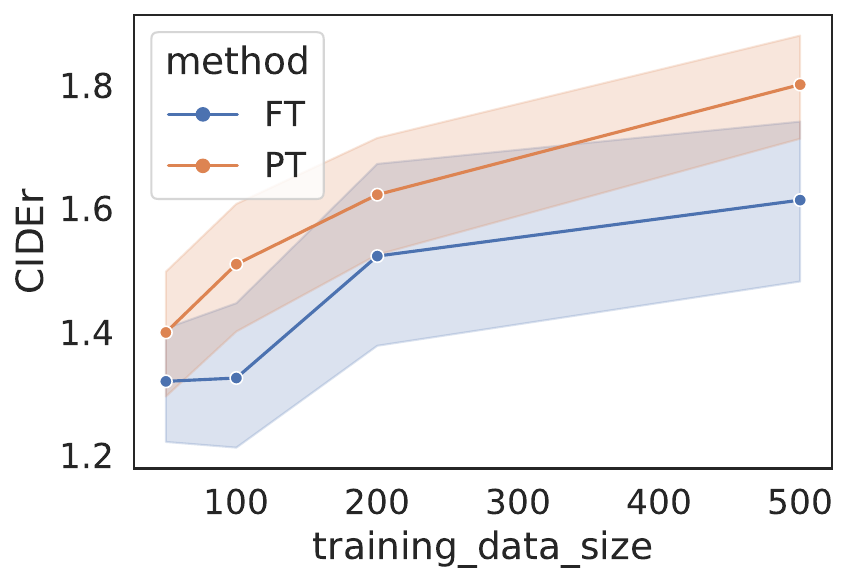} \hspace{-20pt}
\caption{\label{app:lowdata} Prefix-tuning (orange) outperforms fine-tuning (blue) in low-data regimes in addition to requiring many fewer parameters. The top three plots correspond to summarization, measured by ROUGE-1, ROUGE-2, and ROUGE-L. 
The bottom three plots correspond to table-to-text, measured by NIST, METEOR, and CIDEr. The x-axis is the training size and the y-axis is the evaluation metric (higher is better).}
\end{figure*}

\cref{app:lowdata} supplements the low-data performance curves in \cref{tab:lowdata} by plotting the relationship between training size and generation metrics for both prefix-tuning and fine-tuning. 

\subsection{Additional Results for the Initialization Experiment}

\begin{figure*}[]
 \includegraphics[page=2,width=0.45\textwidth]{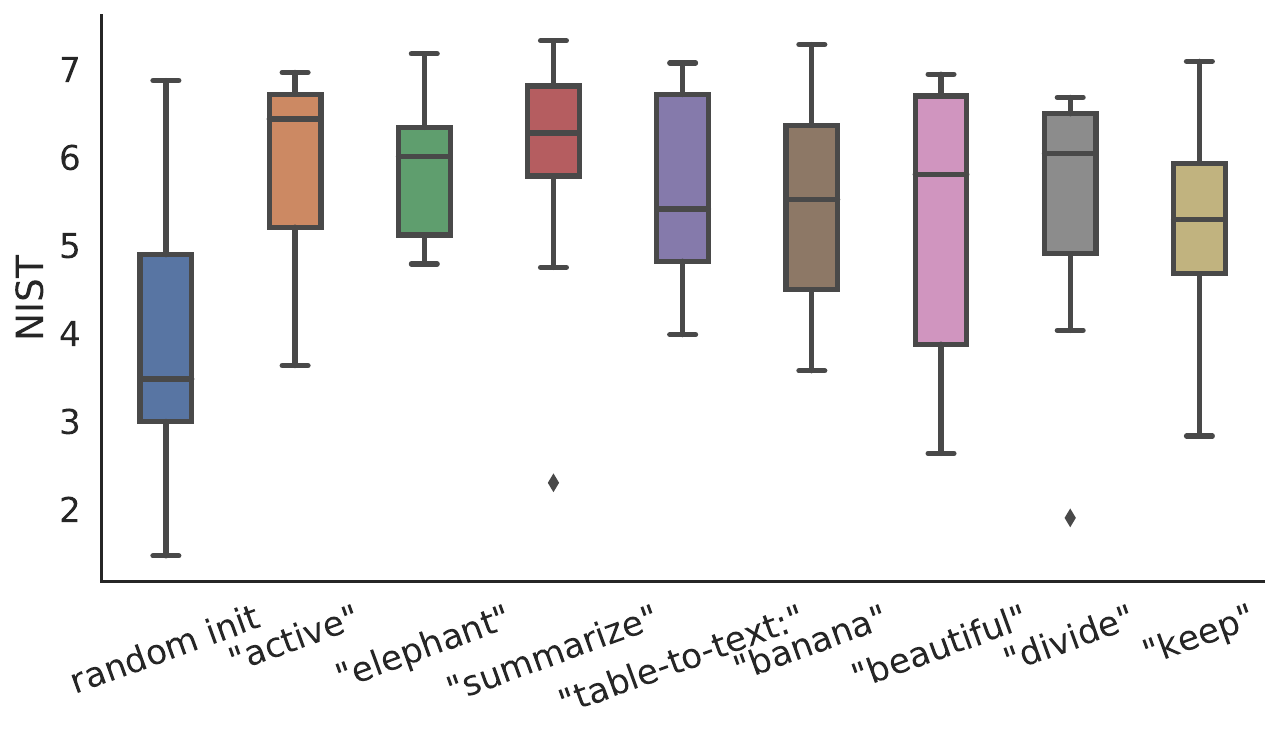}
 \qquad
 \includegraphics[page=4,width=0.45\textwidth]{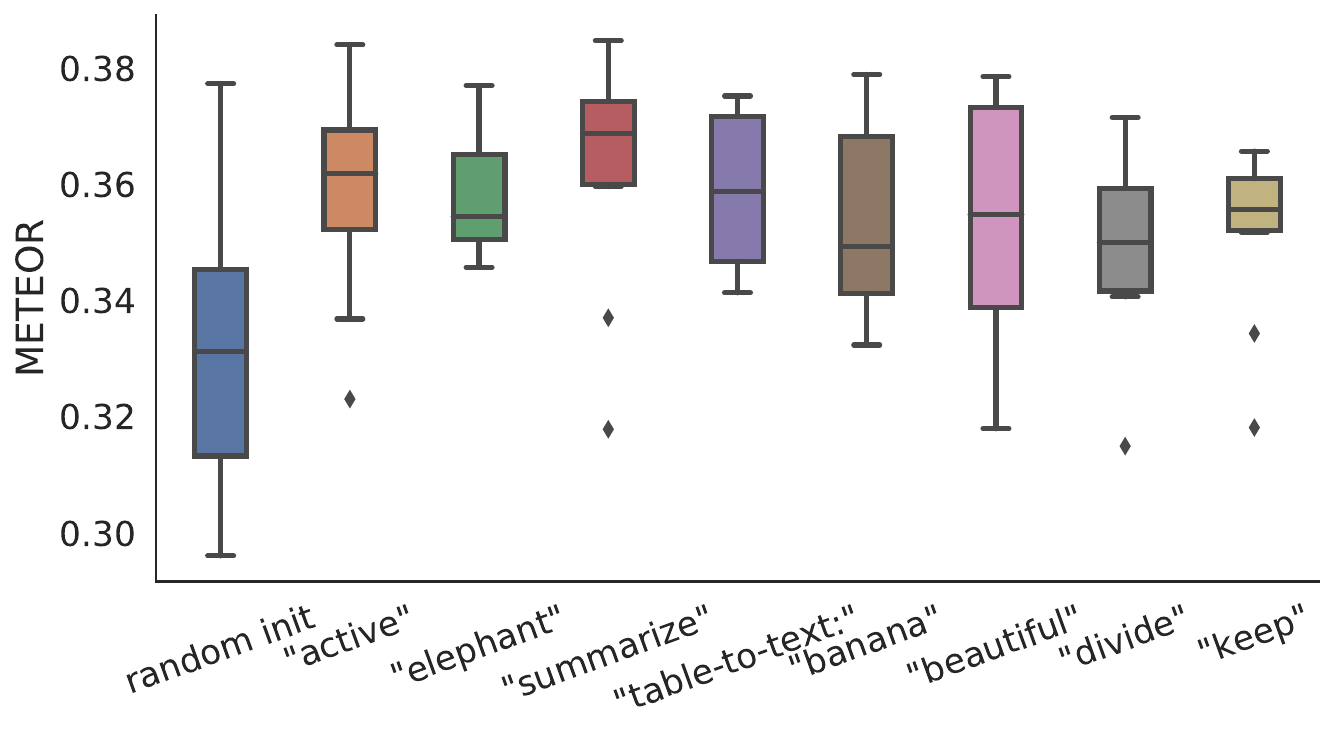}\hspace{-20pt}\\
 \qquad
 \includegraphics[page=3,width=0.45\textwidth]{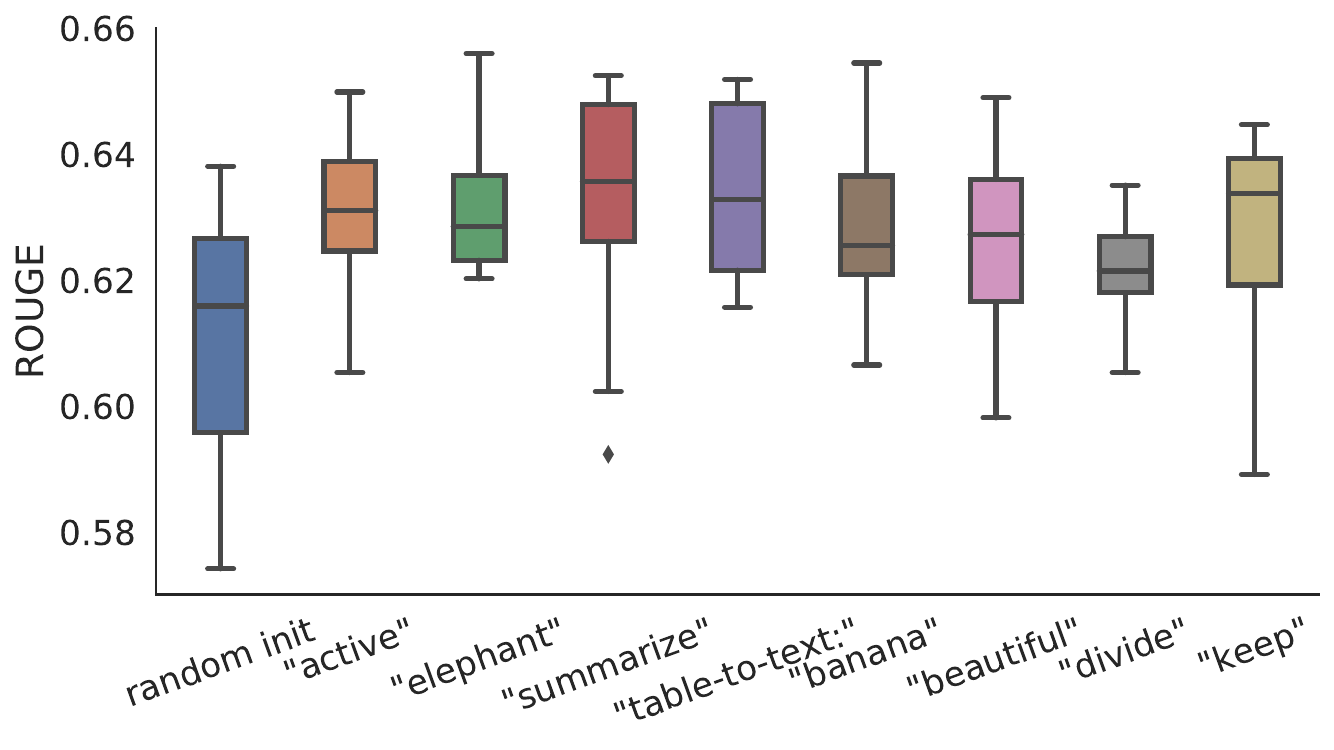}
 \qquad
 \includegraphics[page=5,width=0.45\textwidth]{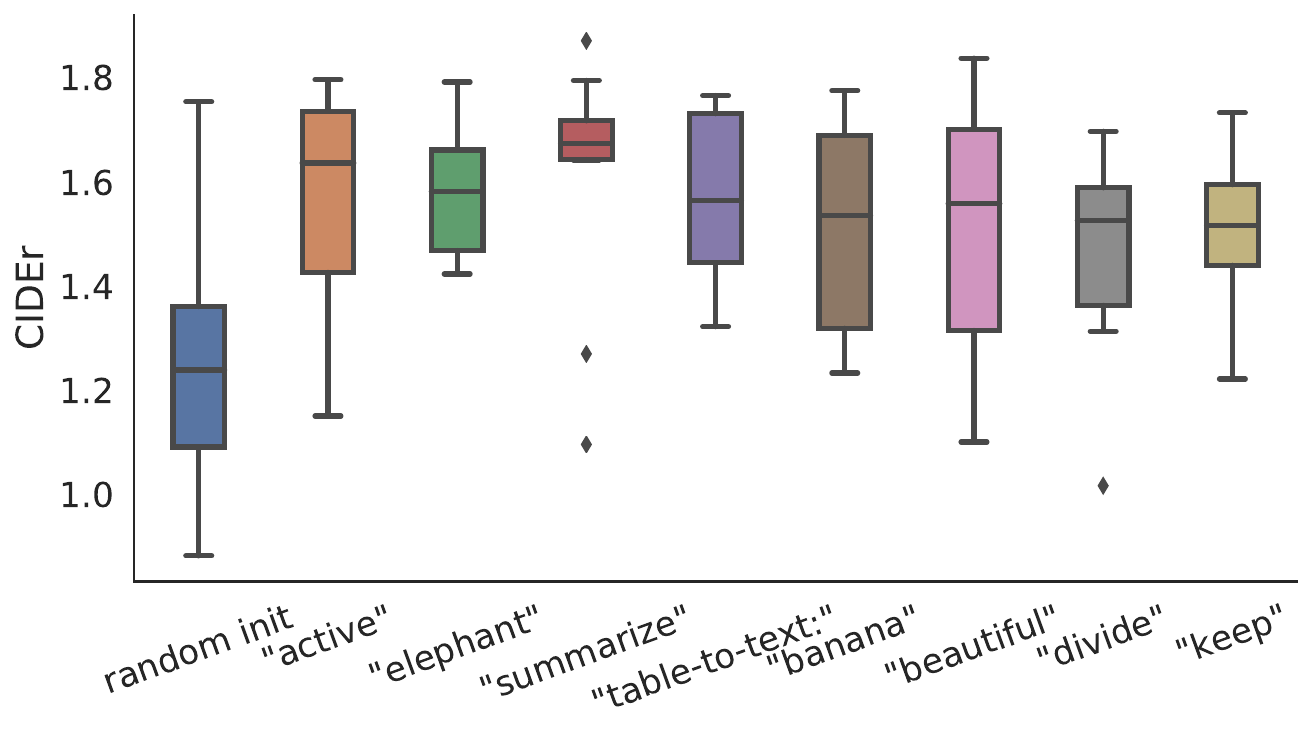}\hspace{-20pt}
 \caption{\label{app:init} Initializing the prefix with activations of real words significantly outperforms random initialization, in a low-data setting with 100 training data. }
\end{figure*}

\cref{app:init} supplements \cref{tab:lowdata} by plotting additional metrics for our initialization technique \cref{ssec:init}. It validates that random initialization (from a uniform (0,1) distirbution) significantly underperforms initializing with real words; Additionally, initializing with task-relevant words (e.g., ``summarization'' and ``table-to-text'') attains slightly better generation scores than initializing with task-irrelevant words (e.g., ``elephant'' and ``banana''). 

\subsection{Qualitative Examples for Extrapolation}
\cref{app:qualitative-webnlg} contains qualitative examples from both seen and unseen categories in WebNLG. We find that for unseen categories, both prefix-tuning and fine-tuning tend to undergenerate (generated output do not cover full table contents) or generate untruthfully  (generated output is inconsistent with table contents). 
In particular, prefix-tuning tends to undergenerate whereas fine-tuning tends to generate untruthfully. 
For seen categories, both perform fairly well in terms of coverage and truthfulness. 

\begin{table*}[]
    \centering
    \resizebox{\linewidth}{!}{
    \begin{tabular}{lp{17cm}}
        Source [Unseen, Athelete] & (Al Kharaitiyat SC, club, Alaa Abdul-Zahra), (Al Khor, ground, Al Kharaitiyat SC), (Shabab Al-Ordon Club, club, Alaa Abdul-Zahra) (Amar Osim,  manager, Al Kharaitiyat SC) \\
        \midrule
        Prefix-tuning & Al Kharaitiyat SC are managed by Amar Osim and play at their ground at Al Khor. Al Kharaitiyat SC are also the club for which Alaa Abdul-Zahra is a player. \\
        Fine-tuning   &  Alaa Abdul-Zahra plays for Al-Kharaitiyat SC and Shabab Al-Ordon Club. He also plays for Al-Khor and manages Al-Kharaitiyat SC.\\
        Reference   & Alaa Abdul Zahra plays for Al Kharaitiyat SC which is located at Al Khor and managed by Amar Osim. The Shabab Al-Ordon club is associated with Alaa Abdul-Zahra.\\
        \toprule
        Source [Unseen, Transportation]  &  (Genoa, location, Costa Crociere), (AIDA Cruises, operator, AIDAstella), (Costa Crociere, owner, AIDAstella)\\
        \midrule
        Prefix-tuning &  AID Astella is operated by Aida Cruises and is owned by the Costa Rican tourist resort of Genoa. \\ 

        Fine-tuning   &  AID Astella, operated by AIDA-Cruises, is located in Genoa and is owned by the Costa Rican government.\\ 
        Reference   & Costa Crociere is the owner of the AIDAstella and are based in Genoa. The operator of AIDAstella is AIDA Cruises. \\ 
 
       \toprule
        Source [Unseen, Politician]  &  (Euro, currency, Netherlands), (Stellendam, birthPlace, Ab Klink ), (Netherlands, nationality, Ab Klink)\\
        \midrule
        Prefix-tuning &  Ab Klink was born in Stellendam and is a national of the Netherlands where the currency is the Euro.\\ 

        Fine-tuning   &  Ab Klink is a national of the Netherlands where the currency is the Euro. He was born in Stellendam.\\ 
        Reference   & Ab Klink was born in Stellendam in the Netherlands, where the national currency is the euro.\\ 
       \toprule
        Source [Unseen, Politician]  &  (Robert E, Lee, commander, Battle of Salem Church), (American Civil War, isPartOfMilitaryConflict, Battle of Salem Church), (Battle of Salem Church, battles, Aaron S. Daggett)\\
        \midrule

        Prefix-tuning &  Robert E. Lee was the commander of the Battle of Salem Church which was part of the military conflict in the American Civil war.\\ 

        Fine-tuning   &  The Battle of Salem Church is part of the American Civil War and was commanded by Robert E. Lee.\\ 
        Reference   & Robert E Lee was a commander in the Battle of Salem Church, which was one of the military conflicts in the American Civil War. Aaron S Daggett fought in the same battle. \\ 

        \toprule
        Source [Unseen, Artist]  &  (Christian alternative rock, musicSubgenre, Alternative rock), (Alternative rock, genre, Andrew White (musician))\\
        \midrule

        Prefix-tuning &  Andrew White is a Christian alternative rock musician.\\ 

        Fine-tuning   &  Andrew White, a Christian alternative rocker, performs.\\ 
        Reference   & The musician Andrew White's genre is alternative rock, the genre which has the sub genre Christian alternative rock.\\ 
        \toprule
        Source [Unseen, Artist]  &  (Hip hop music, genre, Allen Forrest),  (solo singer, background, Allen Forrest) \\
        \midrule

        Prefix-tuning &  Allen Forrest is a solo singer.\\ 

        Fine-tuning   &  Born in\\ 
        Reference   & Allen Forrest is a solo singer whose genre is Hip Hop music.\\ 
        \toprule
        Source [Seen, ComicsCharacter]  &  (Americans, nationality, Ducan Rouleau),  (Ducan Rouleau, creator, Baymax),(Alan Tudyk, starring, Big Hero 6 (film)), (Steven T Segle, creator, Baymax), (Big Hero 6 (film), serires, Baymax) \\
        \midrule

        Prefix-tuning &  Baymax is a character in Big Hero 6 which stars Alan Tudyk. He was created by Steven T. Seagle and the American, Duncan Rouleau.\\
        Fine-tuning   &  Alan Tudyk stars in the film Big Hero 6 in which Baymax is a character created by Steven T. Seagle and the American, Duncan Rouleau.\\ 
        Reference   & Baymax is a character who appeared in Big Hero 6 starring Alan Tudyk. It was created by Steven T Seagle and the American, Duncan Rouleau.\\ 

        \toprule

        Source [Seen, City]  &  (Washington, D.C., capital, United States),  (White Americans, ethnicGroup, United States), (United States, country, New Jersey), (New York City, largest City, United States), (New Jersy, isPartOf, Atlantic City)\\
        \midrule

        Prefix-tuning &  Washington D.C. is the capital of the United States where the largest city is New York City and the White Americans are an ethnic group. Atlantic City, New Jersey is also part of the United States. \\
        Fine-tuning   &  Atlantic City, New Jersey is part of New Jersey in the United States. The capital city is Washington D.C. and one of the ethnic groups is White Americans.\\ 
        Reference   & New York City (NYC) is the largest U.S. city. Atlantic City, New Jersey are also part of the United States with its capital as Washington, DC and home to White Americans.\\

        \bottomrule
    \end{tabular}}
    \caption{\label{app:qualitative-webnlg}  Qualitative examples from WebNLG. The first 6 examples are from the unseen categories, labeled next to source; the last two examples are from the seen categories. 
    For unseen categories, both prefix-tuning and fine-tuning tend to undergenerate (generated output do not cover full table contents) or generate untruthfully  (generated output is inconsistent with table contents). 
    In particular, prefix-tuning tends to undergenerate more often than  generate untruthfully whereas fine-tuning tends to  generate untruthfully. 
    For seen categories, both perform fairly well in terms of coverage and truthfulness. }
\end{table*}

\end{document}